\documentclass{article}

\usepackage[final]{nips_2018}

\usepackage[utf8]{inputenc}
\usepackage[T1]{fontenc}
\usepackage{hyperref}
\usepackage{url}
\usepackage{amsfonts,amssymb,amsthm,amsmath}
\usepackage{microtype}
\usepackage[dvipsnames]{xcolor}
\usepackage[inline]{enumitem}
\usepackage{tikz}
\usetikzlibrary{cd,arrows,calc,shapes,positioning}

\newcommand{\sevendashtxt}{\bgroup%
\sbox0{7}\usebox0\llap{\rule[.5\ht0]{.5\wd0}{.15\ht0}\rule{.24\wd0}{0pt}}\egroup}
\def\examples{\emph{Experiments:}\ }

\newcommand{\cX}{\mathcal{X}}
\newcommand{\cY}{\mathcal{Y}}
\newcommand{\cL}{\mathcal{L}}
\DeclareMathOperator{\supp}{supp}
\renewcommand{\vec}[1]{\mathbf{#1}}

\theoremstyle{definition}
\newtheorem*{observation*}{Observation}
\theoremstyle{plain}
\newtheorem{proposition}{Proposition}
\newtheorem*{corollary*}{Corollary}

\tikzstyle{startstop} = [rectangle,
                         rounded corners,
                         minimum width={\minwidth},
                         minimum height={\minheight},
                         text centered,
                         draw=black,
                         fill=ForestGreen!30]
\tikzstyle{io} = [trapezium,
				          trapezium left angle=70,
                  trapezium right angle=110,
                  minimum width={\minwidth},
                  minimum height={\minheight},
                  text centered,
                  draw=black,
                  fill=MidnightBlue!30]
\tikzstyle{process} = [rectangle,
                       minimum width={\minwidth},
                       minimum height={\minheight},
                       text centered,
                       text width={1.7cm},
                       draw=black,
                       fill=BrickRed!30]
\tikzstyle{decision} = [diamond,
                        minimum width={\minwidth},
                        minimum height={\minheight},
                        text centered,
                        inner sep=1pt,
                        draw=black,
                        aspect=1.5,
                        fill=Orange!50]
\tikzstyle{arrow} = [thick,->,>=latex]

\def\minheight{0.2cm}
\def\minwidth{1cm}

\newcommand{\perfectlearningflowchart}{%
\begin{tikzpicture}[node distance=0.8cm and 0.5cm, auto,font=\small]
\node (start) [startstop] {observed effect $y$};
\node (match) [decision, below= of start] {$f(x) = y$};
\node (causal) [process, left= of match] {causal model $f$};
\node (anticausal) [process, above= of causal, xshift={-0.6cm}] {anti-causal model $f^{-1}$};
\node (next) [io, left= of causal, xshift={0.1cm}, align=center] {next\\cause $x$};
\node (stop) [startstop, right= of match, fill=MidnightBlue!30, xshift={0.3cm}] {output $x$};
\draw [arrow] (start) -- (match);
\draw [arrow] (next) -- (causal);
\draw [arrow] (causal) -- (match);
\draw [arrow] (match) -- node[anchor=south] {yes} (stop);
\draw [arrow] (match) -- ++(0,-1) -| (next) node[pos=0.25,anchor=north] {no};
\draw [arrow, dotted] (start) -- (anticausal);
\draw [arrow, dotted] (anticausal) -| (next);
\draw[dashed] (-5.6,-3.15) rectangle +(6.6,2.4);
\node[anchor=north, black!70] at (-2.4, -3.15) {validation by the causal model};
\end{tikzpicture}}

\setlist[itemize]{noitemsep, topsep=0pt,leftmargin=*}
\setlist[enumerate]{noitemsep, topsep=0pt,leftmargin=*}

\title{Generalization in anti-causal learning}

\author{%
Niki Kilbertus\textsuperscript{1, 2, $\ast$}\\
\texttt{nkilbertus@tue.mpg.de}
\And
Giambattista Parascandolo\textsuperscript{1, 3, $\ast$}\\
\texttt{gparascandolo@tue.mpg.de}
\And
Bernhard Sch\"olkopf\textsuperscript{1, $\ast$}\\
\texttt{bs@tue.mpg.de}
\AND
\normalfont
\normalsize\textsuperscript{1}Max Planck Institute for Intelligent Systems\\
\normalsize\textsuperscript{2}University of Cambridge\\
\normalsize\textsuperscript{3}Max Planck ETH Center for Learning Systems\\
\normalsize\textsuperscript{$\ast$}authors are listed in alphabetical order
}

\begin{document}

\maketitle

\begin{abstract}
The ability to learn and act in novel situations is still a prerogative of animate intelligence, as current machine learning methods mostly fail when moving beyond the standard i.i.d.\ setting.
What is the reason for this discrepancy?
Most machine learning tasks are anti-causal, i.e., we infer causes (labels) from effects (observations).
Typically, in supervised learning we build systems that try to \emph{directly} invert causal mechanisms.
Instead, in this paper we argue that strong generalization capabilities crucially hinge on searching and validating meaningful hypotheses, requiring access to a causal model.
In such a framework, we want to find a cause that leads to the observed effect.
Anti-causal models are used to drive this search, but a causal model is required for validation.
We investigate the fundamental differences between causal and anti-causal tasks, discuss implications for topics ranging from adversarial attacks to disentangling factors of variation, and provide extensive evidence from the literature to substantiate our view.
We advocate for incorporating causal models in supervised learning to shift the paradigm from inference only, to search and validation.
\end{abstract}

\textit{I would rather discover one cause than gain the kingdom of Persia.} --- Democritus

\section{Introduction}
\label{sec:intro}

We have seen impressive successes of supervised learning in a wide variety of applications, usually when large sets of i.i.d.\ data are available, either from human labelling or simulations.
The i.i.d.\ setting is also well understood in theory, but more complex tasks such as transfer learning, domain adaptation or covariate shift are conceptually and theoretically much harder.
In this paper, we argue that those problems benefit from taking into account a causal generative structure.
Here we only consider the two extreme settings of either purely causal or anti-causal learning.
The former refers to a situation where we are learning to predict effect from cause, and for the latter it is vice versa \citep{SchoelkopfICML2012}.
We discuss when and why these tasks should be solvable outside the i.i.d.\ setting.

Statistical learning theory provides a rigorous framework describing when learning is possible \citep{vapnik1998statistical,cucker2002mathematical}.
The typical setting consists of
\begin{enumerate*}[label=\textbf{(C\arabic*)},topsep=0pt]
  \item two measurable spaces $(\cX, \Sigma_{\cX})$ and $(\cY, \Sigma_{\cY})$,
  \item a fixed unknown \emph{sampling distribution} $P(X)$ on $\cX$ (which need not have full support),
  \item a fixed unknown conditional distribution $P(Y|X)$, and
  \item a finite sample~$S$, i.i.d.\ drawn from the joint distribution $P(X, Y) = P(Y|X) P(X)$.
\end{enumerate*}
From~$S$, we then wish to learn a mapping $\cX \to \cY$ that ``performs well'' on new data drawn from $P(X,Y)$.
One way to think of $P(Y|X)$ is that there exists an unknown mapping from $\cX$ to $\cY$ that induces $P(Y|X)$, e.g., a physical law plus measurement noise, in which case we seek to learn an approximation of the law.

Consistency and convergence guarantees, for example within the empirical risk minimization (ERM) framework \citep{vapnik1998statistical}, paired with universal function approximation guarantees for neural nets \citep{hornik1989,cybenko1989} prompt the conventional wisdom that ``with enough data and powerful models \emph{anything can be learned}''.
In the present paper, we use the informal term \emph{weak generalization} to refer to  generalization to unseen data from the same distribution $P(X,Y)$ under the assumptions \textbf{(C1-4)}.
Formally, for ERM this corresponds to the notion of consistency, \citep{vapnik1998statistical,poggio2004general}.
Since the empirical distribution function of~$S$ converges uniformly to the one of $P(X)$ as $n \to \infty$ \citep{glivenko1933sulla,cantelli1933sulla}, i.e., eventually~$S$ covers all of~$\supp(P(X))$, we can view weak generalization as a form of \emph{interpolation}.

In practice, many tasks violate the i.i.d.\ assumption, instead requiring some form of \emph{extrapolation} outside the support of~$P(X)$.
Even though conditions for finite sample generalization bounds on different domains, e.g., \citep{shaiben2010}, are rarely considered, let alone quantified in applications, the ``anything can be learned'' paradigm taught us to maintain high expectations with respect to learnability.
From a conceptual viewpoint, this is just a leap of faith.
However, instead of worst-case bounds, let us take an optimistic viewpoint starting from the following informal observation.

\emph{Law of continuity.} Our world is goverend by universal continuous laws (or mechanisms).\footnote{``Natura non facit saltus'' (nature does not make jumps) is stated as an axiom in works of Gottfried Leibniz, also known as the ``law of continuity''.}

If this holds true, then one could justify the above mentioned leap of faith by the following \emph{intuitive belief}:
our models do not just \emph{imitate} the underlying mechanism on $\supp(P(X))$, but (approximately) \emph{identify} it \citep{vapnik1998statistical}.
Subject to this intuitive belief of identification, we may plausibly hope to generalize to samples outside $\supp(P(X))$ in some cases.
One could call this \emph{strong generalization}.\footnote{Strong generalization here is not related to strong universal consistency in learning theory.}

In section~\ref{sec:causalanticausal} we investigate arguments why and when the intuitive belief may hold true in practice, and discuss its relation to the key differences between causal and anti-causal learning.
Section~\ref{sec:fundamentally} provides examples from various domains where even well-behaved functions often have complex inverses.
Following that, section~\ref{sec:integrating} states that if the causal model is known, a search procedure can ensure strong generalization in the anti-causal direction.
We then explicitly draw connections to various fields of machine learning.

\textbf{Contributions.}
The present paper advocates a paradigm shift regarding generalization, develops insights, and draws connections some of which have not previously been made explicit.
Specifically, we
\begin{enumerate}
  \item argue that in general strong generalization is limited in anti-causal learning and that access to the causal model can remove such limitations;
  \item point out direct consequences for several areas of machine learning, including deep learning and encoder-decoder architectures, adversarial examples, disentangling factors of variation;
  \item discuss exploitation of the causal model for anti-causal learning through search and validation as a general principle (see Figure~\ref{fig:searchandvalidate}), and point out existing experimental evidence supporting this hypothesis in the literature.
\end{enumerate}

\section{Causal and anti-causal learning}
\label{sec:causalanticausal}

In causality, domain robustness of the causal mechanism is captured by the \emph{principle of the independence of cause and mechanism} (ICM) \citep{daniusis2010inferring,SchoelkopfICML2012,shajarisales2015telling}.
We assume that the causal mechanism $P(Y|X)$ is invariant with respect to changes in the distribution of the cause $P(X)$ \citep{pearl2009causality,peters2017elements}, i.e., $P(Y|X)$ is a physical mechanism that applies for \emph{any} input distribution $P(X)$ on $\cX$.
This is not true for the anti-causal direction: in the generic case, $P(X|Y)$ is not invariant under $P(Y)$ when $Y$ is an effect of $X$, see \citep{peters2017elements}.
Statistical learning theory contains this asymmetry as assumption \textbf{(C3)}, i.e., a fixed mechanism $P(Y|X)$, governing the statistical dependence in the joint distribution $P(X,Y) := P(Y|X)P(X)$.
This characterizes the inherent difference between learning the two directions.
From here on we always assume that $X$ causes $Y$ (i.e. $X$ always denotes causes and $Y$ effects), and for simplicity we rule out confounding.\footnote{We note that confounded settings, where both $X$ and $Y$ are effects caused by other variables, share aspects with the anti-causal setting \citep{SchoelkopfICML2012}.}

We can interpret the ICM principle as a statement about feeding different cause distributions to an invariant causal mechanism $P(Y|X)$.
In particular, these distributions could have different support, i.e., $\supp(P(X))$ might be a proper subset of $\cX$, which is then the entire domain on which $P(Y|X)$ can be applied.
This is a typical setup for domain adaptation or tasks with covariate shift.
Thereby, ICM is closely related to the observation that natural rules are universal and that the \emph{law of continuity} holds true in the causal direction.
The intuitive belief implies that the precise (continuous) mathematical structure of causal mechanisms is such that machine learning models efficiently identify them when trying to fit the training data.
\emph{Under the assumption of the ICM principle and the intuitive belief about identification we can expect strong generalization in the causal direction.}

This is a positive message; however, it is unfortunately the case that in many machine learning tasks we seek to learn the anti-causal direction $P(X|Y)$.
Whether one estimates the true parameters of a system yielding certain measurements, infers personal characteristics or sentiments of humans from their resulting behavior, or identifies underlying objects from their visual, acoustic, or textual representations, anti-causal learning is ubiquitous in practice.

Often we do not want to infer the cause from the observed effect in full detail, but are only interested in certain properties or aspects thereof.
For example, in MNIST digits, the cause can be considered a human brain state encoding the intent to write a certain digit (higher level concept of interest).
The causal mechanism is a complex sequence of biochemical reactions and sensorimotor movements, eventually resulting in the hand-written digit (effect) \citep{peters2017elements}.
An abstract class label, say a digit class \emph{7}, is not the actual input to this mechanism, but a macroscopic property subsuming the brain states (low-level causes) intending to write a \emph{7}.
Clearly, this is a highly idealized description of a complex phenomenon.

\begin{figure}[t!]
\centering
\begin{minipage}{0.48\textwidth}
  \centering
  \includegraphics[width=\textwidth]{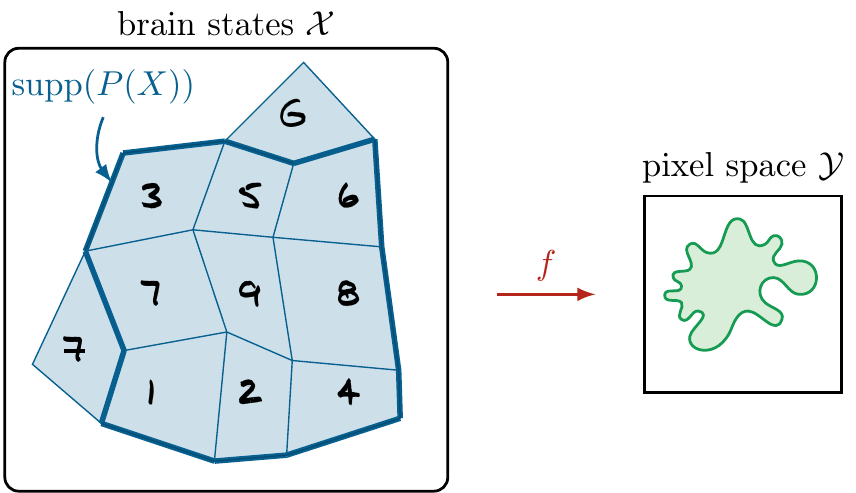}
  \caption{The low-level causal mechanism for MNIST digits and the high-level assigned labels.}
  \label{fig:mnist_example}
\end{minipage}%
\hfill
\begin{minipage}{0.48\textwidth}
  \centering
  \resizebox{\textwidth}{!}{\perfectlearningflowchart{}}
  \caption{A model for the anti-causal direction that implements the causal model and uses it to find the correct cause by a (heuristically guided) exhaustive search.}
  \label{fig:searchandvalidate}
\end{minipage}
\end{figure}

Consequently, the support of $P(X)$ is defined by all causes that constitute the concepts (labels) of interest.
Denoting the set of labels by~$\cL$, we write $S_l \subset \cX$ for the causes belonging to class $l \in \cL$.
Figure \ref{fig:mnist_example} illustrates this setting for the MNIST example, where each numbered cell represents one subset $S_l$.
Note that while we require $\supp(P(X)) \subset \bigcup_{l \in \cL} S_l$---i.e., all possible training examples have a label---one could have $S_l \backslash \supp(P(X)) \ne \emptyset$.
For example, while during training we might only encounter \emph{7}s without a dash, we still assign the same label to \emph{\sevendashtxt{}}s with dashes.

Moreover, there might be causes entirely outside~$\supp(P(X))$ that can still be labeled meaningfully.
In our example, we consider hand-written letters to be produced by the same causal mechanism applied to similarly structured causes, e.g., a brain state intending to draw the letter G.
This is the starting point for various transfer tasks such as continual (e.g., few-shot) learning, where novel classes are introduced dynamically.
We conclude that the macroscopic properties of interest in a given task determine the support of $P(X)$, e.g., the digits determine which brain states we consider admissible causes.
Often, it can naturally be extended to causes that still have meaningful labels, for example slightly altered digits, or letter, which gives rise to our expectations of strong generalization.

While the ICM principle might provide some justification for such expectations in the causal direction, these do not hold for the anti-causal direction.
In the next section we provide examples where well-behaved functions are hard to invert.

\section{Hardness of the anti-causal direction}
\label{sec:fundamentally}

Let us now explore differences between well-behaved functions and their inverses in various domains and examples.
While some are not directly related to machine learning (in which case we do not speak about causality), they share the mathematical description and relevant computational nature.

\textbf{Ill-posed inverse problems.}
\citet{vapnik1998statistical} distinguishes between \emph{imitating} and \emph{identifying} the underlying mechanism $\cX \to \cY$.
While \emph{imitation} is captured by ERM, \emph{identifying} the mechanism from training data typically forms a so called inverse problem, for which only asymptotic results exist and which is generically ill-posed.\footnote{%
A problem is ill-posed, if no unique solutions exists or it is unstable, i.e., little input perturbations lead to vastly different outputs.
As is usual in the theory of inverse problems, we assume existence and uniqueness and only concern ourselves with stability.
In fact, non-injectivity issues do not arise for probabilistic mappings and are typically avoided in linear deterministic inverse problems by working with the Moore-Penrose generalized inverse.}
Over the last decades, inverse problems received increasing attention, as scientists realized that nature is not universally well-behaved.\footnote{\citet{vapnik1998statistical} remarks that
``The theory of ill-posed problems can be considered one of the most important achievements in understanding the nature of many problems.
[\ldots] Hadamard thought that ill-posed problems are purely mathematical phenomenon
and that the real-life problems were well-posed.
Soon, however, it was discovered that there exist important real-life problems that are ill-posed.''}
From the point of view of the present paper, this is no contradiction to the {\em Law of continuity}, but rather a consequence of the fact that we are often looking at anti-causal problems.
Small perturbations of effects, e.g., measurement noise, can then cause arbitrarily large deviations of the inferred cause.
An exemplary mathematical statement is that the Moore-Penrose generalized inverse of a linear bounded operator between Hilbert spaces need not be continuous.

The connections between inverse problems and statistical learning theory (in the weak generalization regime) have recently been discussed by \citep{kurkova2004learning,rosasco2005learning} and shown to have practical consequences \citep{bertero1988ill}.
Thus, even though we can reasonably assume stability for the causal direction, this need not hold for the anti-causal direction.
Interestingly, a certain form of \emph{leave-one-out stability} of the learning algorithm provides necessary and sufficient conditions for generalization in an even more general class of learning problems than ERM \citep{mukherjee2006learning,bousquet2002stability,feldman2018generalization}.
This highlights the close connections between learnability and stability.

\textbf{Causal discovery via minimum description length (MDL).}
\citet{janzing2010causal} argue that the causal direction has a shorter description in terms of Kolmogorov complexity\footnote{The Kolmogorov complexity of an object is the shortest program on a given universal Turing machine that outputs the object, see \citep{li2008p} for an overview.} than the anti-causal one.
Practical approaches based on MDL as a proxy for Kolmogorov complexity can successfully identify causal directions from data \citep{budhathoki2016causal,marx2017telling}, {providing evidence that the anti-causal direction has higher expression complexity}.

\textbf{One-way functions.}
The field of cryptography largely relies on functions that are hard to invert, i.e., the (unproven) existence of so called \emph{one-way} functions.
Those are functions that can be computed in polynomial time, but the probability for any randomized algorithm to find an inverse is negligible, see e.g., \citep{goldreich20016foundations}.
Candidates for one-way functions include prime factorization, discrete logarithm, or quadratic residue.
They provide evidence that \emph{inverses of well-behaved functions indeed can have larger execution complexity}.
Specifically, this means that the best known solution to finding a pre-image of a given value is as expensive as an exhaustive search over all inputs.

\textbf{NP problems.}
Similar difficulties are found in NP problems, where it is easy (polynomial time) to verify a given solution, but no efficient algorithm is known to find one in the first place.
The \emph{subset sum problem}, finding a binary vector $\pmb{\alpha} \in \{0,1\}^d$ such that for a known vector $\vec{x} \in \mathbb{R}^d$ and a target $s \in \mathbb{R}$, $\vec{x}^T \pmb{\alpha} = s$, is NP-complete.
Learning, expressing, and executing the forward direction, i.e., the dot product $\vec{x}^T \pmb{\alpha}$, is easy.
Once this operation is identified, it yields consistent results regardless of the values of $\vec{x}, \pmb{\alpha}$.
However, no polynomial time solution for the inverse mapping $(\vec{x}, s) \mapsto \pmb{\alpha} \in \{0,1\}^d$ is known.
Sparse reconstruction under the $\ell_0$ norm, an ill-posed inverse problem relevant to image processing, is essentially a subset-sum problem and thus NP-complete.

\textbf{Restrictions of the input domain.}
Various instances of hard-to-invert mappings share the characteristic that the domain of the forward direction has a natural extension.
For example, multiplication can be defined on all integers (or even larger spaces), in which case~$n \times 1$ is a trivial factorization of~$n$.
However, once we restrict the input domain to prime tuples, factorization (i.e., the inverse) becomes less trivial.
This might be no surprise, because the input domain of the forward direction is just the solution space for the inverse problem, i.e., restrictions shrink the solution space.
Note the resemblance to our observations about $\supp(P(X)) \subsetneq \cX$ in section~\ref{sec:causalanticausal}.

\textbf{Expressing inverses mathematically.}
The Abel-Ruffini theorem states that general polynomial equations of degree greater than 4 do not have an algebraic solution, i.e., one that can be found by a fixed finite sequence of algebraic computations \citep{abel1826demonstration}.
Similarly, some bijective elementary transcendental functions, such as $f(x) = x + \sin(x)$, do not have an elementary inverse.

As argued above, there is ample evidence that inversion can be hard even if the forward mechanisms are simple.
One may then wonder \emph{why does anti-causal learning often work so well in practice?}

First of all, while some anti-causal mappings are ill-behaved, not all are.
At times, they are just as easily expressed, learned, and computed as their causal counterparts.
Moreover, recovering only high-level properties of the original causes can render anti-causal learning easy, e.g., inferring digit labels rather than brain activations from MNIST images.
Most importantly, in the weak generalization regime, theoretical approximation guarantees can cover complex inverse functions.
However, in the present work we are mostly interested in the more challenging case of strong generalization: even the best image classifiers can be fooled by tiny input perturbations and struggle to deal with small domain shifts.
In the next section, we explain how integrating a causal model is sufficient for strong generalization in the anti-causal direction by searching over the space of causes and validating the resulting effect.

\section{Integrating the causal model for anti-casual learning}
\label{sec:integrating}

Unlike for the causal direction, ICM does not justify expectations of strong generalization when directly estimating a function from effects to causes.
We now discuss how to extend our expectations of strong generalization to the anti-causal direction, by exploiting the causal model.
This is in line with a large body of work utilizing generative modeling (without referring to causality and ICM), e.g., \citep{storkey2009training}.

\subsection{Searching Through The Space Of Causes}
\label{sec:formal}

Several of the problems enumerated in section~\ref{sec:fundamentally} can be solved through an exhaustive search.
We now formalize this simple and inefficient but highly effective procedure, enabling strong generalization through iterative search through the space of causes.
Intuitively, under the strong assumption that the causal model is known and may be queried indefinitely, given any observed effect $y$, we can reliably find a viable cause $x$ by simply searching over all causes $\cX$, see Figure \ref{fig:searchandvalidate}.
Access to the causal model is thus a sufficient condition for strong generalization in the anti-causal direction.

\begin{proposition}[Constructive anti-causal learning]
Assume query access to a deterministic causal model $f: \cX \to \cY$ on a finite domain $\cX$.
Then in the anti-causal direction, for each $y \in \cY$ we can determine whether it has a valid cause, and if so, find a cause $x \in \cX$ such that $y$ is the effect of $x$.
\label{th:constructive}
\end{proposition}
This can be done by simply enumerating the elements in $\cX$ and checking whether $f(x)$ equals $y$ until either a match is found or there are no $x\in \cX$ left.
While mathematically this procedure directly extends to countably infinite $\cX$, it may not halt in finite time.
Hence, in the general case of infinite $\cX$, the interesting part is to find a good strategy for selecting inputs, i.e., smart search heuristics, that either find a match quickly, or have high confidence that there does not exist a viable cause after a small number of checks.

For example, to classify an image of a cat, we could run a photo-realistic 3D render engine (causal mechanism) over a rich library of objects from all angles (causes) until we reproduce the image, in which case we return the name of the rendered object.
In practice, heuristics or techniques like dynamic programming for certain NP problems can guide the search over causes, which often allows for efficient approximate solutions.
The problem of learning the anti-causal direction given the causal model thus reduces to finding an efficient search strategy.
The structure in the anti-causal direction and our priors determine how efficiently we can perform this search.
Since in the i.i.d.\ setting supervised machine learning models often capture this structure surprisingly well, they are prime candidates for efficient heuristics to drive the search over the space of causes.

We compare this to optimization, where highly structured objectives (e.g., convex ones) allow for deterministic optimization schemes with convergence guarantees.
Weaker forms of regularity may still allow for procedures that work well in practice (various Monte Carlo methods).
However, for an arbitrary complex mapping, one cannot avoid exhaustive search over all possible inputs.
In optimization too, constraints typically impede guiding heuristics, as proposed causes must now reside in the allowed domain.

Note that Proposition~\ref{th:constructive} assumes that the causal model and its entire domain of applicability are known.
In practice, if the causal model is not known already, one would have to learn it too, which might be hard in itself.
The crucial aspect is that the causal direction will be a well-behaved function as argued in section~\ref{sec:intro}, which in principle can be expected to generalize across domains.

\subsection{Implications For Machine Learning}
\label{sec:consequences}

We now point out practical consequences of the difference between causal and anti-causal learning.
Further, we suggest that the idea of incorporating causal models in anti-causal tasks constitutes a unifying interpretation of advances in diverse applications.
Experimental results addressing this hypothesis are found in the literature and we point to evidence substantiating our view.

\paragraph{Deep Learning.}
Deep learning is motivated by the prevailing ``anything can be learned'' mindset, and supported by the ICM principle in the causal direction even for strong generalization, see section~\ref{sec:causalanticausal}.
However, it does not strongly generalize in the anti-causal direction.
To implement a search as in Proposition~\ref{th:constructive} (see also Figure \ref{fig:searchandvalidate}) in a practical deep learning architecture, a model that can adapt the number of computational steps is required.
While Turing complete models such as recurrent neural networks (RNN) \citep{siegelmann1995computational} can \emph{in principle} implement such a search, feed forward networks such as CNNs and MLPs perform a constant (or upper bounded) number of computational steps for any input.
Such \emph{constant computation models} can not implement constructive anti-causal learning.

\examples \citet{eslami2016attend,putzky2017recurrent} use an RNN to iteratively refine inference over a known generative model of the data.
This allows such a model to identify more objects at once at test time than it was presented with during training, a sign for strong generalization \citep{eslami2016attend}.
Recent works analyzing how to adaptively control the computation time of RNNs include \citep{graves2016adaptive,figurnov2017probabilistic}.

\paragraph{Adversarial Examples.}
State-of-the-art models, especially in deep learning, are vulnerable to \emph{adversarial examples}: slightly perturbed or wildly out-of-distribution examples that lead to erroneous predictions with high confidence, even for models that achieve impressive weak generalization as measured by i.i.d.\ test error \citep{szegedy2013intriguing}.
When dropping the intuitive belief about identification, i.e., losing robustness, such failure modes are to be expected for anti-causal learning.
One can view adversarial examples as failure to generalize strongly.
To the best of our knowledge, adversarial examples have only been demonstrated in the anti-causal direction with a single exception for a generative variational auto-encoder (VAE) architecture \citep{kos2017adversarial}.
In the paragraph below on encoder-decoder architectures, we argue that this still constitutes an attack on an anti-causal component, namely on the encoder.
In principle, constructive anti-causal learning should be robust to adversarial examples.
\begin{corollary*}
Constructive anti-causal learning as specified in Proposition~\ref{th:constructive} cannot be fooled.
\end{corollary*}
This is an immediate consequence of the fact that constructive anti-causal learning determines whether a given $y \in \cY$ has a valid cause, and---if it does---identifies one.
Because a given cause has a single label and leads to one effect only, the procedure never assigns a wrong label to~$y$.
While the render engine approach to classification is likely inefficient for random search, it is impossible to fool.
Rendering \emph{any} tea pot will never produce a picture similar to one of a cat with small added noise unless imperfections of the renderer can be exploited.
If all possible causes are exhausted without a match, e.g., if the image was extensively manipulated, it does not have a valid cause anymore and it would (correctly) not be classified.

\examples \citet{wang2017safer,samangouei2018defensegan,song2017pixeldefend,ilyas2017robust} increase robustness against adversarial examples, by learning a generative (causal) model and synthesizing images until a match for the input is found.
Main challenges for this approach remain learning reliable causal models and developing efficient search heuristics.

\paragraph{Encoder - Decoder.}
The goal of autoencoders is often to learn a meaningful and small hidden representation of the data containing all necessary information to reconstruct it.
Recent work argues that decoders and encoders correspond to causal and anti-causal mappings respectively \citep{weichwald2014causal,BesSunSch18}.
We argue that for strong generalization, the latent representation should correspond to the causes of the input data, the decoder should implement the causal mechanism, and the encoder should implement its inverse, i.e., estimate underlying causes from observations.
In this case, the decoder may be expected to generalize across tasks.
During training it verifies the estimated causes, acting as a regularizer.
On the other hand, there is no reason to expect the encoder to generalize strongly.

\examples \citet{besserve2017group,BesSunSch18} study properties of learned causal generative models that distinguish them from anti-causal models such as encoders or discriminators.
The adversarial attacks on a seemingly causal generative model in \citep{kos2017adversarial}, can be viewed as attacks on the encoder part, i.e., again on an anti-causal task.
We will encounter further limitations of the encoder in VAEs throughout the following discussion about disentangling factors of variation.

\paragraph{Disentangling Factors Of Variations.}
Disentangling independent factors of variations---despite receiving increasing attention by the community---still lacks an agreed-upon definition of what independent factors of variations are.\footnote{This was highlighted by the organizers of the dedicated 2017 NIPS Workshop on \emph{Learning Disentangled Representations} \url{https://sites.google.com/view/disentanglenips2017} as a main open question.}
Authors usually rely on the ``naturalness'' of such factors, which lends itself to a causal interpretation: a set of independent causes (the factors) are entangled through a set of causal mechanisms that generate the effects (the data we observe).
Consequently, the task of disentangling factors of variations translates to the anti-causal task of producing a set of independent causes from the observed effects.

Based on the previous paragraph, it makes sense that most state-of-the-art methods for disentangling are based on autoencoder architectures \citep{higgins2016beta,kim2018disentangling,kumar2017variational}.
Even though the encoder (anti-causal part) would in principle suffice to recover the causes, adding the decoder corresponds to integrating a causal model, which reconstructs the effects and \emph{verifies} the estimated latent code through the associated loss.
If the decoder were the true causal model, no estimated latent representation that is not a valid cause would pass this verification.

However, while a feed forward decoder should suffice for the causal direction, a constant computation encoder inherently limits the strong generalization capabilities, because it cannot iterate on the decoder.
An adaptive encoder with access to an accurate causal model could in principle implement the exhaustive search for constructive anti-causal learning.

Disentangling consists of two independent tasks.
The first is to discover good latent dimensions, i.e., identifying the domain of the causal mechanism.
The second is to assign a specific point in this latent space to each example, i.e., inference in the anti-causal direction.
Even if the causal mechanism and its input domain are known, the remaining second task, disentangling, is still hard for any given input.
State-of-the-art approaches focus on the first task, largely disregarding the difficulty of the second.

\examples \citet{lake2015human} achieve stronger generalization by learning a causal generative model of handwritten characters based on composition of strokes using probabilistic programming, and iteratively sampling from it.
\citet{desjardins2012disentangling} successfully disentangle factors through inference over a (causal) generative model of the data.

\paragraph{Bayesian Inference.}
Often, the prior in Bayesian inference captures our beliefs about properties of the world underlying some collected data.
From this perspective, it can be viewed as a distribution over possible causes.
The likelihood $P(Y|X)$ then may represent the causal mechanism: given a cause $X$, how likely is it that nature produces the effect $Y$.
The object of interest is the posterior $P(X|Y)$, i.e., a distribution over possible causes given that effect $Y$ has been observed.
Bayes' rule enables us to express the anti-causal direction as $P(X|Y) = \frac{P(Y|X)P(X)}{P(Y)}$. We can read this as follows: given $Y$, sample causes from $P(X)$ and apply the forward mechanism $P(Y|X)$. If it produces $Y$ (with high probability), $X$ is a good candidate cause.
Thus, Bayesian inference implements anti-causal learning as in Proposition~\ref{th:constructive} with a built-in search procedure, namely to sample from the prior over causes.
Randomized searches like Monte Carlo sampling techniques are among the most common approaches to Bayesian inference.

We provide examples on parameter estimation and approximate Bayesian computation (ABC) in the next paragraph on ``Incorporating known underlying causal models''.
It has been shown that approximating probabilistic inference and finding maximum a posteriori probabilities in Bayesian belief networks or learning large-scale discrete Bayesian networks are all NP-hard \citep{dagum1993approximating,shimony1994finding,chickering2004large}, consistent with our interpretation (see ``NP problems'' in section~\ref{sec:fundamentally}).

\paragraph{Incorporating Known Causal Models.}
When causal models or reliable approximations thereof are known in advance, they can readily be used for validation in the anti-causal direction.

\examples In natural sciences causal models are often solvable analytically or numerically with high accuracy, facilitating the forward search.
For example, to estimate the masses and spins of binary black-hole systems from their generated gravitational waves \citep{abbott2016properties}, the waveforms for hundreds of thousands parameter settings sampled via MCMC are simulated and then matched with the recorded data.
Simulation of an (approximately) known causal model also lies at the heart of ABC \citep{rubin1984bayesianly}, a method for likelihood-free Bayesian inference of the posterior that has proven potent in many areas, see \citep{beaumont2002approximate,toni2009approximate,beaumont2010approximate}.
In computer vision, 3D render engines have successfully been used in a line of research known as \emph{vision as inverse graphics} to invert the physical process that produced an image \citep{loper2014opendr,eslami2016attend}.
\citet{lake2013one} learn a causal model of image generation to outperform the state-of-the-art in one-shot classification. \citet{tassa2012synthesis} solve a set of MuJoCo tasks in real time \emph{without any learning}, by online planning over the simulator itself.
AlphaGo \citep{silver2016mastering} evaluates moves by searching on the real causal model of the game, guided by a network trained to recognize patterns from experts' games to generate candidate moves.
Similarly, recent approaches to program induction from input-output pairs assist various search techniques with a deep neural net trained to predict certain properties of the program (anti-causal) \citep{balog2017deepcoder}.

\paragraph{Connections To Animate Intelligence.}
It has been conjectured that humans internally build powerful causal models for simulations and approximate physical mechanisms through intuitive theories \citep{krynski2007role,waldmann2006beyond,hamrick2011internal,battaglia2013simulation,gerstenberg2017intuitive}.
The fact that we dream and imagine constitutes further evidence that we use causal generative models.
Because we rely on these robust models in anti-causal tasks \citep{longcamp2003visual,james2006letter}, we are quick to infer causal relations even when there are none \citep{pronin2006everyday,matute2015illusions}.

Richard Feynman describes the process of discovering (or learning) physical laws as follows \citep[Chapter 7]{feynman}:
\begin{enumerate*}[label=(\roman*)]
\item Guess a theory,
\item compute its consequences, and
\item compare them to nature.
\end{enumerate*}
If the predictions do not match experimental results, repeat.
This description of the scientific method is akin to constructive anti-causal learning.
A theory in cognitive psychology suggests that humans induce causal relations from observation by domain-general statistical inference (search) that is guided by domain-specific prior knowledge (heuristics) \citep{griffiths2009theory}.
Further, \citet{ullman2010theory} directly describe \emph{theory acquisition as a stochastic search}.
Analogously, scientists do not guess theories entirely at random, but have strong intuition and guiding principles.

Finally, we draw connections to the two modes of reasoning---``fast System 1'' and ``slow System 2''---as presented in \citep{kahneman2011thinking}.
While System 1 is associative, domain specific, and independent of the working memory, System 2 is rule based, domain general, and limited by working memory.
Comparing these attributes to our arguments, we suggest implementing specialized anti-causal models akin to System 1 that are known to generalize weakly, as well as slow iterative search over causal models akin to System 2 that step in for unfamiliar settings where extrapolation is key.

\section{Related work}

The study of causality and its role in statistics and machine learning has a long history and
recently causal reasoning has been claimed vital for solving future challenges in AI \citep{lake2017building,pearl2018theoretical}.
Cognitive psychologists and neuroscientists have acknowledged that humans seem to exploit causal models to learn in novel anti-causal tasks.
While the difference between causal and anti-causal with respect to semi-supervised learning and strong generalization tasks (e.g., covariate shift) have been discussed on an abstract level \citep{SchoelkopfICML2012}, we are not aware of an explicit account of the direct consequences for current state-of-the-art methods on popular tasks and the unifying underlying interpretation we presented.

\section{Conclusion}

There are key differences between learning in the causal and anti-causal direction in terms of extrapolation or strong generalization capabilities.
We have argued that strong generalization in the anti-causal direction is often difficult, and how it benefits from integrating a causal model: anti-causal models should drive the search and causal models should be used for validation and extrapolation.
We have discussed a number of implications of our findings for various machine learning tasks---including deep learning and encoder-decoder architectures, adversarial examples, disentangling factors of variation---providing evidence in support of our claims by reference to experimental and theoretical results in the literature.
We hope that a paradigm shift of supervised learning from direct inference to search and validation by incorporating causal models in anti-causal tasks will ultimately lead to improved robustness and generalization of machine learning systems.

\bibliographystyle{apalike}
\bibliography{refs}

\end{document}